\documentclass[pdflatex,sn-mathphys-num]{sn-jnl}

\usepackage{colortbl}
\usepackage{float}
\usepackage{adjustbox}
\usepackage[utf8]{inputenc}
\usepackage[T1]{fontenc}
\usepackage{hyperref}       
\usepackage{url}
\usepackage{booktabs}       
\usepackage{amsfonts}
\usepackage{nicefrac}       
\usepackage{microtype}      
\usepackage{xcolor}         

\usepackage{algorithm}
\usepackage{algpseudocode}
\usepackage{amsmath}
\usepackage{algorithmicx}
\usepackage{array}
\usepackage{textcomp}
\usepackage{stfloats}
\usepackage{verbatim}
\usepackage{graphicx}
\usepackage{enumitem}

\usepackage{mathtools}
\usepackage{amsthm}
\usepackage{subcaption} 
\usepackage{multirow}
\usepackage{pifont}
\newcommand{\ccy}[1]{\cellcolor{yellow!30}#1}
\newcommand{\cco}[1]{\cellcolor{orange!25}#1}
\newcommand{\ccr}[1]{\cellcolor{red!25}#1}

\theoremstyle{thmstyleone}%
\theoremstyle{thmstyletwo}%

\theoremstyle{thmstylethree}%

\begin{document}

\title[Article Title]{Multi-Sample Anti-Aliasing and Constrained Optimization for 3D Gaussian Splatting}


\author[1]{\fnm{Zheng} \sur{Zhou}}\email{m320123332@sues.edu.cn}
\equalcont{These authors contributed equally to this work.}

\author[1]{\fnm{Jia-Chen} \sur{Zhang}}\email{m325123603@sues.edu.cn}
\equalcont{These authors contributed equally to this work.}

\author*[1]{\fnm{Yu-Jie} \sur{Xiong}}\email{xiong@sues.edu.cn}
\author[1]{\fnm{Chun-Ming} \sur{Xia}}\email{cmxia@sues.edu.cn}

\affil*[1]{\orgdiv{School of Electronic and Electrical Engineering}, \orgname{Shanghai University of Engineering Science}, \orgaddress{\city{Shanghai}, \postcode{201620},  \country{China}}}


\abstract{Recent advances in 3D Gaussian splatting have significantly improved real-time novel view synthesis, yet insufficient geometric constraints during scene optimization often result in blurred reconstructions of fine-grained details, particularly in regions with high-frequency textures and sharp discontinuities. To address this, we propose a comprehensive optimization framework integrating multisample anti-aliasing (MSAA) with dual geometric constraints. Our system computes pixel colors through adaptive blending of quadruple subsamples, effectively reducing aliasing artifacts in high-frequency components. The framework introduces two constraints: (a) an adaptive weighting strategy that prioritizes under-reconstructed regions through dynamic gradient analysis, and (b) gradient differential constraints enforcing geometric regularization at object boundaries. This targeted optimization enables the model to allocate computational resources preferentially to critical regions requiring refinement while maintaining global consistency. Extensive experimental evaluations across multiple benchmarks demonstrate that our method achieves state-of-the-art performance in detail preservation, particularly in preserving high-frequency textures and sharp discontinuities, while maintaining real-time rendering efficiency. Quantitative metrics and perceptual studies confirm statistically significant improvements over baseline approaches in both structural similarity (SSIM) and perceptual quality (LPIPS).}




\maketitle

\begin{figure*}[ht]
	\centering
	\includegraphics[width=\textwidth]{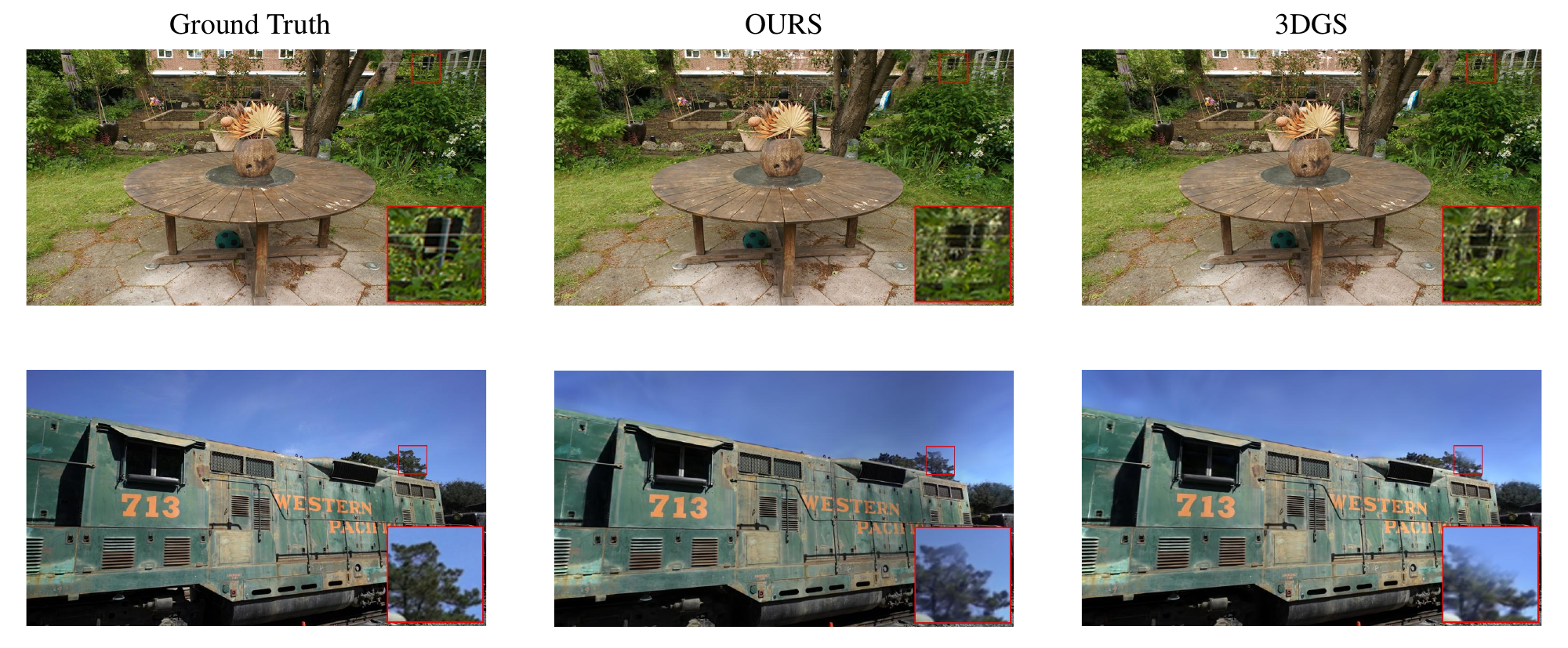}
	\caption{The difference between rendered images and real images. In comparison to the first line of images, the window rendered by 3DGS loses its geometric shape and the edges become unclear. The comparison of the second line image shows that the 3DGS rendering result is missing some content.}
	\label{contrast}
\end{figure*}

\section{Introduction}\label{sec1}
Novel view synthesis (NVS) has emerged as a cornerstone capability in 3D computer vision~\cite{shum2000review}, enabling photorealistic image generation from arbitrary viewpoints—a critical requirement for applications spanning virtual reality, computational cinematography, and immersive media. Conventional approaches reconstruct explicit 3D scene geometry from multi-view inputs to ensure view consistency, but often face fidelity limitations under complex geometric or illumination conditions~\cite{seitz2006comparison}.
The advent of Neural Radiance Fields (NeRF)~\cite{mildenhall2021nerf} marked a paradigm shift by coupling implicit scene representations with differentiable volume rendering, achieving unprecedented view synthesis quality. Subsequent variants addressed efficiency concerns through strategies like cone tracing~\cite{barron2021mip,barron2022mip}, hash-grid encodings~\cite{muller2022instant}, and sparse volumetric representations~\cite{fridovich2022plenoxels,garbin2021fastnerf}. Nevertheless, these methods frequently sacrifice high-frequency detail when accelerating rendering, particularly in high-resolution regimes.
3D Gaussian Splatting (3DGS)~\cite{kerbl20233d} presents a compelling alternative by explicitly modeling scenes as anisotropic 3D Gaussians with adaptive density control. Unlike NeRF's computationally intensive volumetric integration, 3DGS employs efficient splat-based projective rendering, achieving real-time performance while preserving geometric fidelity. However, as illustrated in Figure~\ref{contrast}, 3DGS exhibits pronounced limitations in regions with sparse observational data, manifesting as local detail degradation and blurring artifacts due to inadequate sampling density~\cite{kerbl20233d}.

To address this issue, we present a novel 3DGS framework integrating multi-sampling anti-aliasing~\cite{jimenez2011filtering,akeley1993reality} and edge aware constraints~\cite{xu2015deep,hua2014edge} to reduce rendering blur and enrich local details through multi sampling and specific constraints. Usually, 3D Gaussian Splatting is rendered only through the Gaussian ellipsoid corresponding to the pixel center, which can result in stiff transitions and color jumps~\cite{mai2024ever}. We use 4 sub-pixel points for multi sampling mixed rendering. Combining the information of sub-pixel points for comprehensive rendering will result in smoother color transitions. At the same time, considering the influence of different Gaussian ellipsoid coverage areas, the multi sampling rendering method can focus on the color details and render the image more realistically. In the loss function, we introduce adaptive weights~\cite{johnson2016perceptual} based on the size of the error to address the discrepancy between the predicted results and the actual image. In this way, when the error of certain pixels or regions is large, their influence during the training process will be amplified, making the network pay more attention to these weak reconstruction or difficult to predict regions, thereby avoiding the phenomenon of excessive smoothness or texture loss in local details. In order to enhance the reconstruction effect of details such as object edges and high-frequency textures, we additionally add gradient difference loss on top of the original loss~\cite{zhang2018learning,lim2017enhanced}. By comparing the difference between the predicted image and the real image in the gradient domain, the model focuses more on the local gradient information of the image. Compared to simple L1/L2~\cite{zhao2017loss} or SSIM loss~\cite{wang2004image}, gradient difference constraint can better capture high-frequency textures and structures in images, resulting in sharper and richer visual details. 
The contributions of this work can be summarized in three key aspects.
\begin{itemize}
	\item A 4×MSAA rasterization pipeline for 3DGS that reduces aliasing artifacts through subpixel-aware Gaussian blending.
	
	\item A hybrid loss function combining error-adaptive weighting and gradient difference constraints to enhance detail reconstruction.

    \item Comprehensive benchmarking demonstrating state-of-the-art performance in both quantitative metrics and perceptual evaluations across multiple datasets.
\end{itemize}

\section{Related Work}\label{sec2}
\subsection{Explicit Scene Representations}
Early approaches to novel view synthesis often relied on convolutional neural networks (CNNs)~\cite{flynn2016deepstereo} to fuse and warp multi-view inputs for generating unseen viewpoints. These methods typically compute pixel- or feature-level blending weights that combine source images into approximate target views~\cite{hedman2018deepblending}. While they perform well for moderate viewpoint changes, they frequently struggle with more complex geometry and larger viewpoint shifts, leading to artifacts and blurred details.
Subsequent research shifted toward volumetric ray-marching and scene representations stored in 3D grids. DeepVoxels~\cite{sitzmann2019deepvoxels}, for instance, introduced a persistent voxel-based representation in which rays sample 3D features to synthesize novel views, resulting in improved spatial consistency compared to 2D image-based blends~\cite{niemeyer2020differentiable}. However, these explicit volumetric methods can become computationally and memory intensive at higher resolutions, limiting their scalability in large-scale or high-fidelity scenarios~\cite{martin2021nerfsynthetics}.
\subsection{Implicit Neural Representations}
To address these limitations, research pivoted to NeRF ~\cite{mildenhall2021nerf}, which represents scenes implicitly via multilayer perceptrons (MLPs). By predicting density and color at any continuous 3D coordinate, NeRF achieves photorealistic novel views with strong multi-view consistency. Numerous NeRF extensions~\cite{barron2021mip} have since been proposed to handle challenges such as moving objects, unknown camera poses, few-shot settings, and anti-aliasing~\cite{martin2021nerfsynthetics}. Nevertheless, traditional NeRF-based pipelines often demand extensive training times, and achieving real-time rendering remains a challenge~\cite{reiser2021kilonerf}. Existing acceleration strategies—including smaller specialized MLPs, tensor factorizations, hashing-based encodings, and various compression techniques~\cite{chen2022tensorf}—help reduce these bottlenecks but can diminish high-frequency detail, particularly at higher resolutions~\cite{zhang2020nerfgan}. Despite progress, implicit methods face fundamental challenges in training speed and real-time rendering, motivating hybrid representation paradigms.
\subsection{Hybrid Gaussian Representations}
In pursuit of an improved balance among rendering speed, fidelity, and data efficiency, 3D Gaussian Splatting has emerged as a promising alternative. This approach employs an explicit representation of anisotropic Gaussians in 3D space, modeling local geometry and color through Gaussian parameters optimized end-to-end~\cite{kerbl20233d,linsen2001gaussian}. Unlike methods that rely on large voxel grids or fully implicit functions, 3D Gaussian Splatting utilizes a differentiable splatting process to project these Gaussians onto the image plane, enabling faster training and real-time rendering while preserving high-fidelity detail.
Mip-splatting further addresses aliasing in 3D Gaussian Splatting by introducing a multi-scale representation analogous to mipmapping~\cite{yu2024mip}. Rather than rendering a single level of Gaussians for the entire scene, this technique organizes Gaussian splats into different resolution layers. During rendering, the pipeline adaptively selects the most appropriate layer based on distance and viewing conditions, thereby mitigating aliasing and improving overall rendering efficiency.
Further developments in Multi-scale 3D Gaussian Splatting introduce a multi-resolution strategy to mitigate aliasing in splatting-based rendering~\cite{yan2024multi}. By representing geometry and color information through layered Gaussian splats at different scales, this technique adaptively selects the most appropriate resolution level based on viewing distance and detail requirements, thereby reducing artifacts caused by under- or over-sampling. This approach achieves high-fidelity, anti-aliased results while maintaining computational efficiency.
Scaffold-GS offers a structured framework for arranging Gaussian splats in a hierarchical or grid-based fashion~\cite{lu2024scaffold}. By embedding geometric and color information into a scaffold-like structure, the method adaptively selects the optimal level of detail for each viewing condition, effectively balancing visual fidelity with computational overhead.
Despite these advantages, 3D Gaussian Splatting can exhibit over-reconstruction when excessive or redundant Gaussians accumulate in certain regions. Such overlapping Gaussians may lead to blurring and artifacts, particularly near high-contrast edges or finely detailed textures~\cite{deng2022depthnerf,zhang2020nerfgan}. 
\subsection{Detail Preservation in Neural Rendering}
The pursuit of high-frequency detail preservation in neural rendering has spawned diverse technical approaches, each addressing specific aspects of the aliasing-detail tradeoff. Traditional graphics-inspired anti-aliasing techniques like MSAA~\cite{molnar1994sorting} and temporal anti-aliasing (TAA)~\cite{akenine2019real} achieve subpixel smoothing through supersampling or frame accumulation, yet their direct application to differentiable rendering pipelines remains challenging due to gradient computation constraints. Meanwhile, frequency-domain optimization strategies have emerged as complementary solutions, with methods like FreGS~\cite{zhang2024fregs} performing coarse-to-fine Gaussian densification by exploiting low-to-high frequency components that can be easily extracted with low-pass and high-pass filters in the Fourier space. Despite these advances, current solutions predominantly address aliasing artifacts and detail erosion as separate challenges—MSAA variants focus on signal-space smoothing while neglecting texture sharpness, whereas gradient-based constraints improve local contrast but struggle with subpixel discontinuities. This bifurcation stems from the inherent difficulty in jointly optimizing geometric stability and spectral fidelity through unified loss landscapes, a gap our method bridges through dual-domain regularization.

To address these issues, we propose a method that combines MSAA, an adaptive weighting strategy, and gradient difference constraints. Specifically, we integrate MSAA into the rasterization process to reduce jagged edges and preserve high-frequency details, introduce a pixel-wise weighting mechanism to emphasize regions with larger reconstruction errors, and employ gradient-aware losses to sharpen edges and textures. By simultaneously targeting aliasing and subtle detail restoration, our approach enhances local fidelity while preserving the real-time advantages of 3D Gaussian Splatting.

\begin{figure*}[ht]
	\centering
	\includegraphics[width=\textwidth]{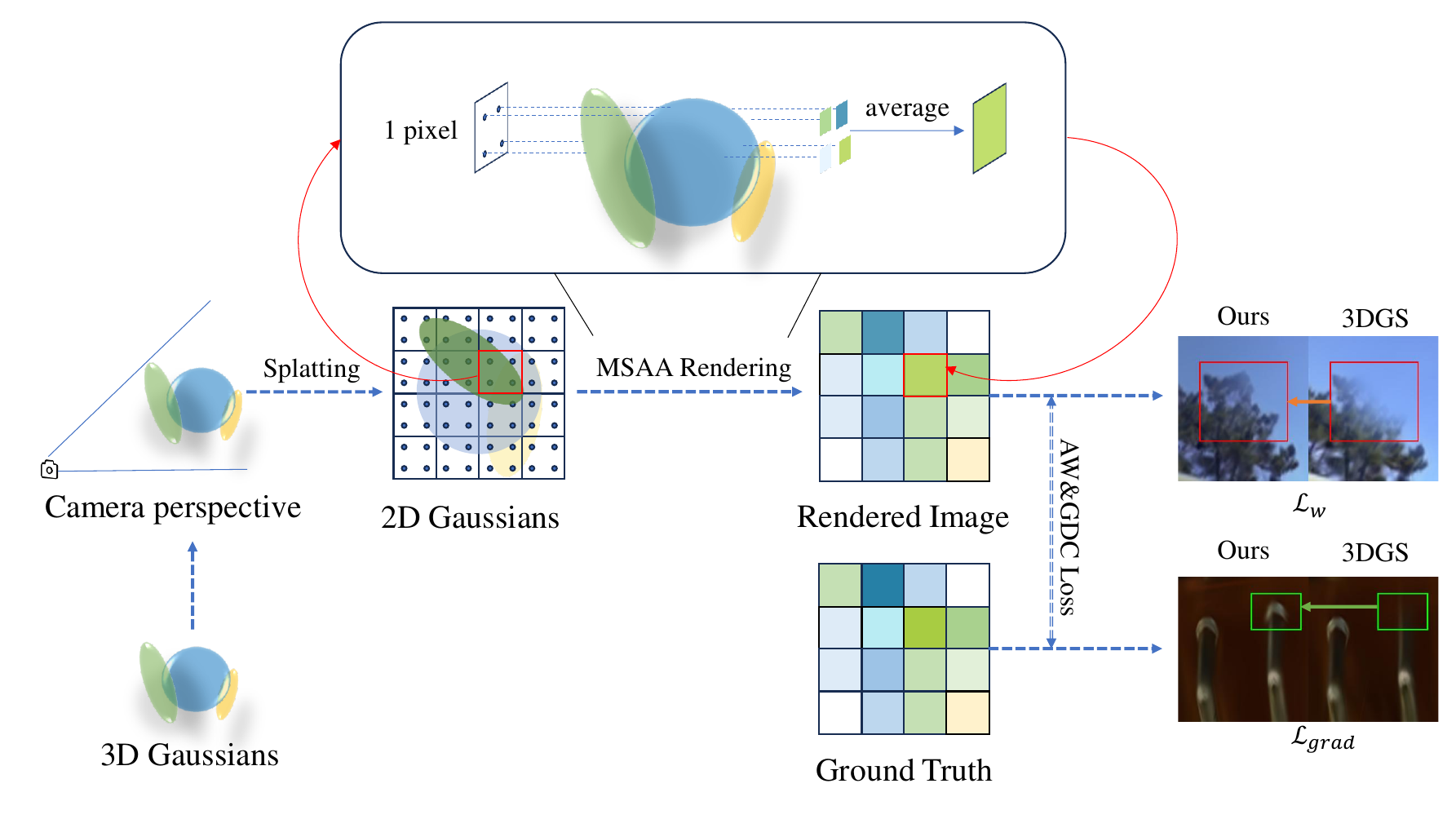}
	\caption{Overview of the Proposed Method. Our approach begins by placing four offset sampling points for each pixel during the rasterization stage. Each sampling point independently performs alpha blending to render the scene, and their resulting colors are averaged to produce the final color for each pixel. This process effectively reduces jagged edges while enhancing the model’s ability to capture both high- and low-frequency details. During backpropagation, we introduce two complementary constraints: an adaptive weighting strategy and gradient difference constraints. The adaptive weighting strategy targets regions suffering from insufficient reconstruction, while the gradient difference constraints address missing boundary details. By guiding the model to focus on these unclear rendering areas, our method ultimately refines the overall rendering quality.}
	\label{overview}
\end{figure*}

\section{Method}
In this section, we first recap the 3D Gaussian Splatting pipeline, including how 3D Gaussians are parameterized and projected onto the image plane. We then describe our proposed enhancements, consisting of three major components: 

\begin{enumerate}
    \item \textbf{Multi-Sample Anti-Aliasing}, which reduces aliasing and jagged boundaries in the rendered images.
    \item \textbf{Adaptive Weighting Strategy}, which selectively increases the training focus on pixels or regions with higher reconstruction error.
    \item \textbf{Gradient Difference Constraints}, which encourage the preservation of high-frequency details and sharp edges.
\end{enumerate}
Together, these techniques address the major shortcomings of naive splatting, notably the loss of fine details and aliasing artifacts. The complete pipeline is shown in the figure \ref{overview}.

\subsection{Preliminaries: 3D Gaussian Splatting}
\label{sec:preliminaries}
We assume our scene is represented by $ N $ anisotropic 3D Gaussians, denoted as $ \{G_n\}_{n=1}^{N} $. Each Gaussian $ G_n $ is described by a center $ \boldsymbol{\mu}_n \in \mathbb{R}^3 $, a covariance matrix $ \Sigma_n \in \mathbb{R}^{3 \times 3} $ which can be decomposed as $ \Sigma_n = R_n S_n R_n^\top $ where $ R_n $ is a rotation matrix and $ S_n $ is a diagonal scaling matrix, a color $ \mathbf{c}_n \in \mathbb{R}^3 $, and an opacity $ \alpha_n \in [0, 1] $.
This explicit representation allows each Gaussian to be learned and updated through gradient-based optimization.

\subsubsection{Projection and Splatting.}
Let $R_{\text{cam}}, \mathbf{t}_{\text{cam}}$ denote the camera extrinsic parameters, and $K$ be the intrinsic matrix. A 3D point $\mathbf{x} \in \mathbb{R}^3$ is mapped onto the image plane via the pinhole model:
\begin{equation}
    \mathbf{p} = \pi\bigl(R_{\text{cam}} \mathbf{x} + \mathbf{t}_{\text{cam}},\, K\bigr) \in \mathbb{R}^2,
\end{equation}
where $\pi(\cdot)$ applies the standard perspective projection using $K$. For each Gaussian $G_n$, its center $\boldsymbol{\mu}_n$ is projected to $\mathbf{p}_n \in \mathbb{R}^2$. The covariance $\Sigma_n^{2D}$ of the elliptical 2D footprint is derived from the original 3D covariance $\Sigma_n$, taking into account the camera projection geometry.

\subsubsection{Color Accumulation.}
At a 2D pixel coordinate $\mathbf{u} \in \mathbb{R}^2$, the contribution of the $n$-th Gaussian can be approximated by a Gaussian weight:
\begin{equation}\label{eq:gaussian_splat}
    w_{n}(\mathbf{u}) \;=\; \exp\!\Bigl(-\tfrac{1}{2}\,
    (\mathbf{u}-\mathbf{p}_n)^\top \,\bigl(\Sigma_n^{2D}\bigr)^{-1}\, 
    (\mathbf{u}-\mathbf{p}_n)\Bigr).
\end{equation}
In practice, the color at $\mathbf{u}$ is computed through weighted or alpha compositing of all Gaussians overlapping that pixel:
\begin{equation}\label{eq:alpha_compositing}
    \mathbf{C}(\mathbf{u}) \;=\; 
    \frac{\displaystyle \sum_{n=1}^{N} \bigl(\alpha_n \, w_{n}(\mathbf{u})\bigr)\,\mathbf{c}_n}
         {\displaystyle \sum_{m=1}^{N} \bigl(\alpha_m \, w_{m}(\mathbf{u})\bigr)}.
\end{equation}
Equations~\eqref{eq:gaussian_splat}--\eqref{eq:alpha_compositing} define the forward splatting step.

\begin{figure}[ht]
	\centering
	\includegraphics[width=\linewidth]{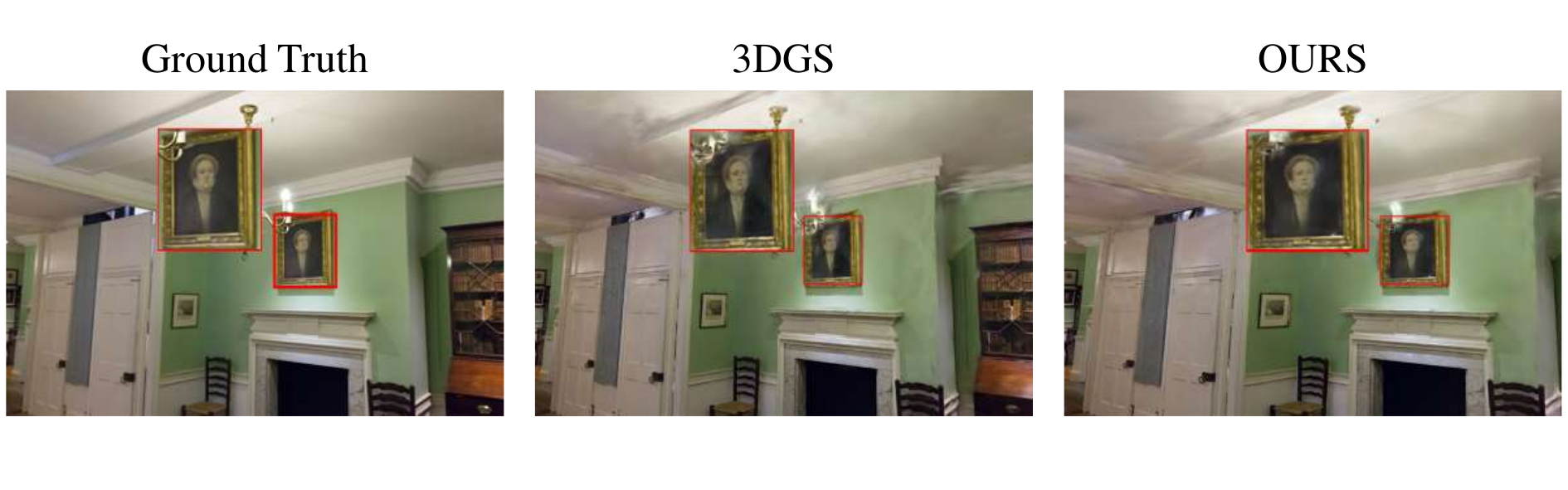}
	\caption{Insufficient detail and blurring issues are observed in the vanilla 3DGS. In particular, the chandelier and frame in the rendered scenes display noticeable distortion and blur, including warped wall surfaces and edges. These artifacts indicate that 3DGS struggles to handle jagged edges and boundaries while maintaining overall scene geometry, resulting in distorted and blurry final images.}
	\label{MSAA1}
\end{figure}

\subsection{Multi-Sample Anti-Aliasing}

While forward splatting maintains a continuous approximation of scene geometry, it introduces aliasing artifacts along high-curvature edges and thin structures, primarily manifested as blurring or jagged discontinuities in reconstructed boundaries, see Figure~\ref{MSAA1}. This phenomenon underscores the inherent limitations of 3DGS in preserving sharp geometric discontinuities and linear feature fidelity. To address these artifacts, we adapt principles from multisample anti-aliasing to the splatting paradigm by extending per-pixel evaluations to a quadruple subsampling scheme. Specifically, our method computes color and geometry contributions across four strategically offset subpixel positions, effectively supersampling edge regions to mitigate high-frequency aliasing. This approach enforces smoother gradient transitions while retaining computational efficiency through optimized subsample weighting, bridging the gap between geometric continuity and edge sharpness in splat-based rendering.

\subsubsection{Subpixel Sampling} 
Let $n\in \mathbb{Z} ^+$ denote the number of subpixel samples per pixel. To mitigate aliasing artifacts while preserving computational tractability, we define a set of subpixel offsets $\{\boldsymbol{\delta}_k\}_{k=1}^{n}$ within the unit square $[0,1]\times[0,1]$, where each $\delta_k=(\delta_{k,x},\delta_{k,y})$represents a fractional displacement from the pixel center. For each pixel center $\mathbf{u}$ on the image grid, the subpixel sampling coordinates are computed as:
\begin{equation}
    \mathbf{u}_k \;=\; \mathbf{u} + \boldsymbol{\delta}_k, 
    \quad k = 1, 2, \dots, n.
\end{equation}
At each subpixel $\mathbf{u}_k$, we perform alpha compositing as defined in Eq.~\eqref{eq:alpha_compositing}, accumulating color contributions from overlapping Gaussians to yield the subsampled color $\mathbf{C}_k(\mathbf{u})$. This stratified sampling strategy effectively captures high-frequency geometric and photometric variations that would otherwise be lost in single-sample-per-pixel rasterization.

\subsubsection{Differentiable Multisample Aggregation} 
The final anti-aliased pixel color $\mathbf{C}^\text{MSAA}(\mathbf{u})$ is computed via a normalized summation over all subpixel samples:
\begin{equation}
    \mathbf{C}^\text{MSAA}(\mathbf{u}) \;=\; 
    \frac{1}{n} \sum_{k=1}^{n} \mathbf{C}_k(\mathbf{u}).
\end{equation}
Critically, this aggregation operator remains fully differentiable, enabling gradient backpropagation through all $n$ sampling paths. The chain rule decomposes the gradient $\partial C^{MSAA}/\partial \Theta $, where $\Theta$ denotes Gaussian parameters. This property encourages the optimization process to prioritize Gaussians whose spatial and spectral properties exhibit view-consistent behavior across subpixel perturbations, thereby enhancing geometric stability and texture fidelity.

\begin{figure}[ht]
	\centering
	\includegraphics[width=\linewidth]{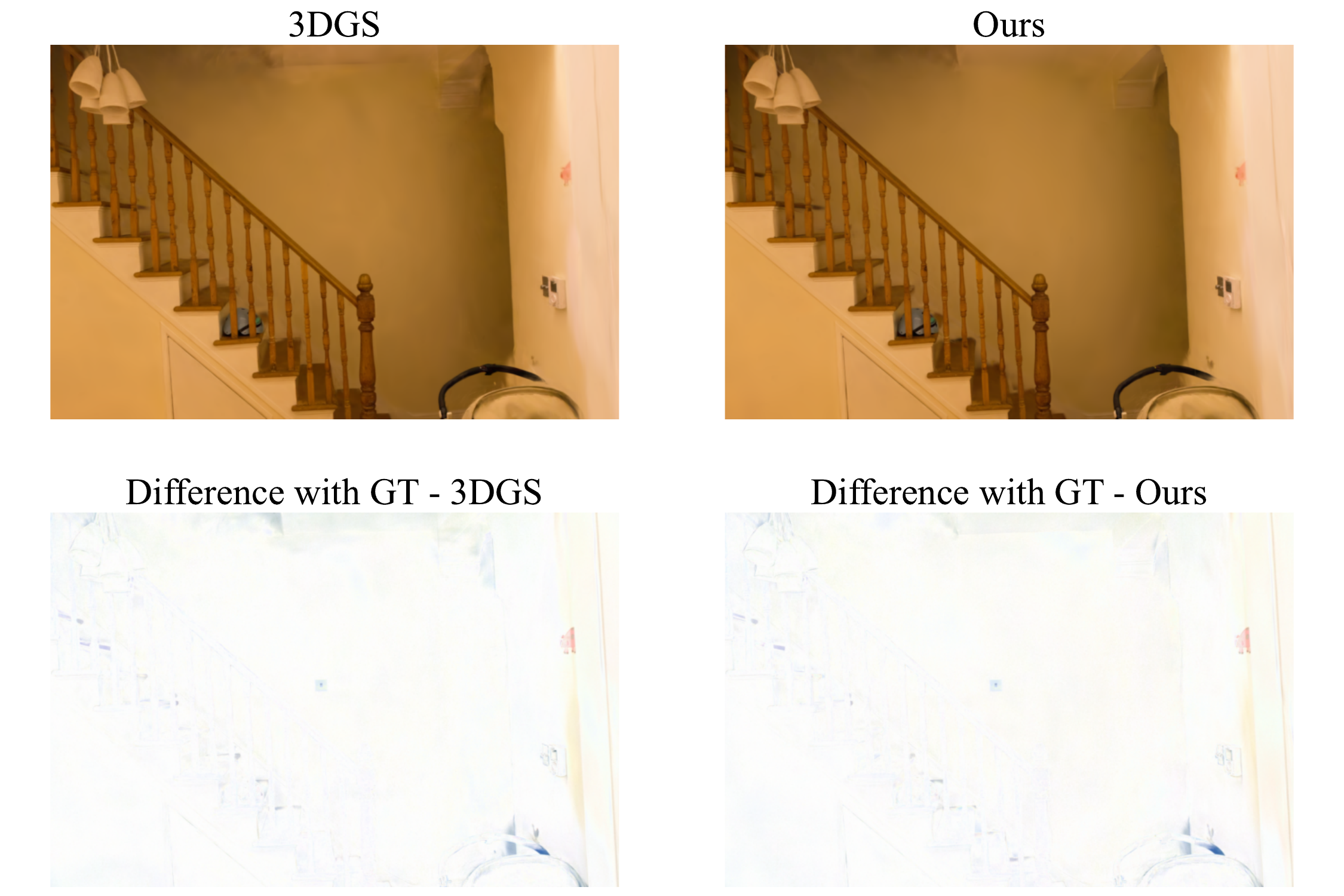}
	\caption{The rendering result on unclear areas. 3DGS may experience insufficient reconstruction when dealing with unclear areas, resulting in severe dissonance and blurring when rendering that area. Our method assigns more weights to areas with insufficient reconstruction, which effectively enhances the reconstruction effect of the area, adds details to unclear areas, and produces more realistic rendering results.}
	\label{AW}
\end{figure}

\subsection{Adaptive Weighting Strategy}

While MSAA effectively mitigates aliasing artifacts, residual reconstruction errors persist in high-frequency regions as shown in Figure~\ref{AW}. Through comparative analysis of our model's outputs against 3DGS renderings, we generate inverse-color difference maps relative to ground truth - where whiter regions indicate closer alignment with reference data. These visualizations reveal significant reconstruction deficiencies in geometrically complex areas such as ceilings and staircases, where 3DGS exhibits substantial detail loss and structural discontinuities.
To address these limitations, we implement an adaptive weighting mechanism within our loss function that dynamically prioritizes pixels exhibiting elevated reconstruction errors. This strategic emphasis enables our model to concentrate learning capacity on regions where conventional rendering approaches struggle, particularly in dense geometric configurations requiring high-frequency detail preservation.

\subsubsection{Pixel-Wise Weight Definition}
Let $I_\text{pred}$ and $I_\text{gt}$ be the predicted and ground-truth images of size $H \times W$. For each pixel $(i,j)$, we first compute the $\ell_1$ error:
\begin{equation}
    e_{i,j} \;=\; \Bigl\lVert I_\text{pred}(i,j) \;-\; 
    I_\text{gt}(i,j) \Bigr\rVert_1.
\end{equation}
To prioritize under-reconstructed regions, we design an adaptive weight map:
\begin{equation}\label{eq:weight}
    w_{i,j} \;=\; \alpha \;+\; (1 - \alpha)\,\frac{\,e_{i,j}\,}
    {\displaystyle \max_{(p,q)\in\Omega } e_{p,q} + \varepsilon},
\end{equation}
where $\Omega=\{1,...,H\}\times\{1,...,W\}$ denotes the spatial domain, $\alpha \in [0,1]$ governs the baseline weight floor, and $\varepsilon>0$ ensures numerical stability. This formulation ensures $w_{i,j}\in [\alpha,1)$ with monotonic increase relative to local error magnitude.

\subsubsection{Weighted Reconstruction Loss}

The error-adaptive weights modulate our reconstruction objective:
\begin{align}
    \mathcal{L}_{\text{w}} 
    &= \frac{1}{\,H W\,} \sum_{i=1}^{H} \sum_{j=1}^{W} 
    w_{i,j}\,\Bigl\lVert I_\text{pred}(i,j) - I_\text{gt}(i,j)
    \Bigr\rVert_1. 
\end{align}

Complementing this, we incorporate a structural regularization term through a decomposed SSIM metric $\mathcal{L}_{\text{D-SSIM}}$, yielding the composite loss:
\begin{align}
    \mathcal{L}_{\text{recon}}
    &= \lambda_1 \,\mathcal{L}_{\text{w}}
    \;+\; \lambda_2 \,\mathcal{L}_{\text{D-SSIM}}.
\end{align}

Here, $\lambda_1$ and $\lambda_2$ control the relative influence of weighted $\ell_1$ and SSIM. By selectively amplifying the loss in high-error regions. This selective error amplification mechanism particularly benefits high-frequency detail recovery where conventional $\ell_1$ losses underperform.

\begin{figure}[ht]
	\centering
	\includegraphics[width=\linewidth]{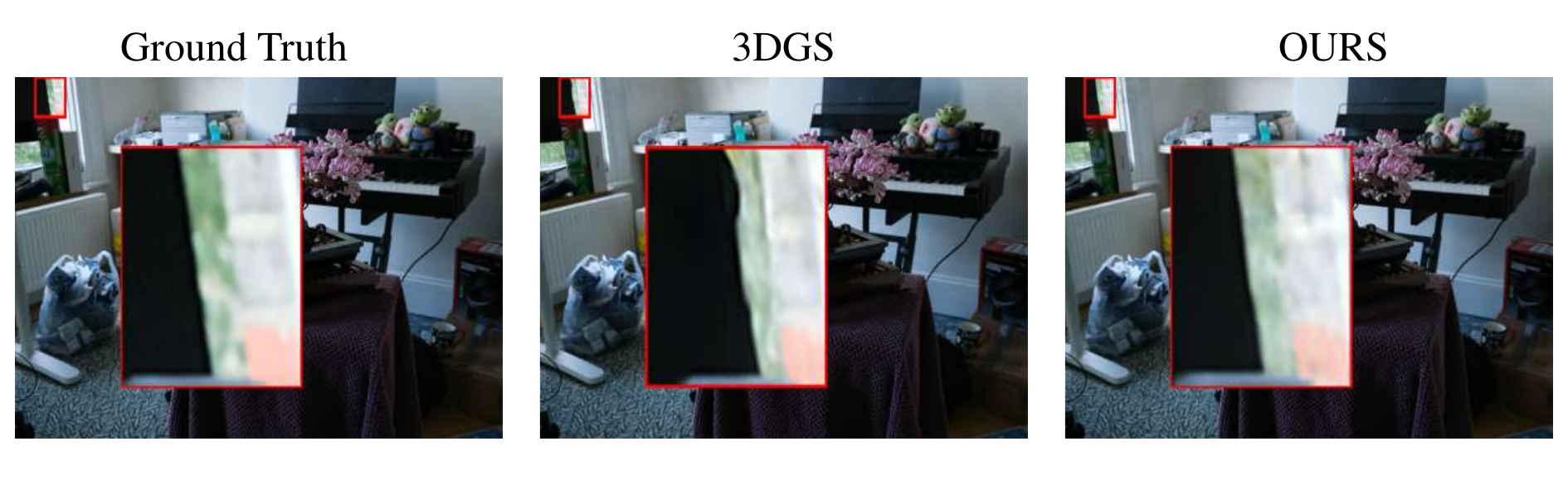}
	\caption{3DGS boundary area reconstruction error, reconstructed straight line boundary as curved boundary. 3DGS does not have effective constraints on boundary information, which results in insufficient attention given to boundary reconstruction, making it difficult to accurately reconstruct boundaries.}
	\label{GDC}
\end{figure}

\subsection{Gradient Difference Constraints}

Although the adaptive weighting mechanism effectively prioritizes high-error regions, it lacks explicit enforcement of edge coherence and high-frequency fidelity – a limitation exemplified in Figure~\ref{GDC} where pixel-wise losses induce excessive smoothing of textural details. To address this critical gap, we introduce a Gradient Difference Constraint (GDC) that directly regulates first-order image variations, ensuring geometric fidelity in boundary reconstruction.

\subsubsection{Discrete Gradient Operators}
For discrete image domains $\Omega=\{1,...,H\}\times\{1,...,W\}$, we define forward-difference operators capturing horizontal and vertical intensity variations:
\begin{align}\label{eq:grad_pred}
    \nabla_x I_\text{pred}(i,j) &= I_\text{pred}(i, j+1) - I_\text{pred}(i, j), \\
    \nabla_y I_\text{pred}(i,j) &= I_\text{pred}(i+1, j) - I_\text{pred}(i, j).
\end{align}
This constructs gradient vector fields for both predicted and ground-truth images:
\begin{align}
    \mathbf{G}_{i,j}^\text{pred} 
    \;&=\; \bigl(\nabla_x I_\text{pred}(i,j),\, 
                 \nabla_y I_\text{pred}(i,j)\bigr), 
    \\
    \mathbf{G}_{i,j}^\text{gt}
    \;&=\; \bigl(\nabla_x I_\text{gt}(i,j),\, 
                 \nabla_y I_\text{gt}(i,j)\bigr).
\end{align}

\subsubsection{Multi-Scale Gradient Alignment}
\begin{figure*}[ht]
	\centering
	\includegraphics[width=\textwidth]{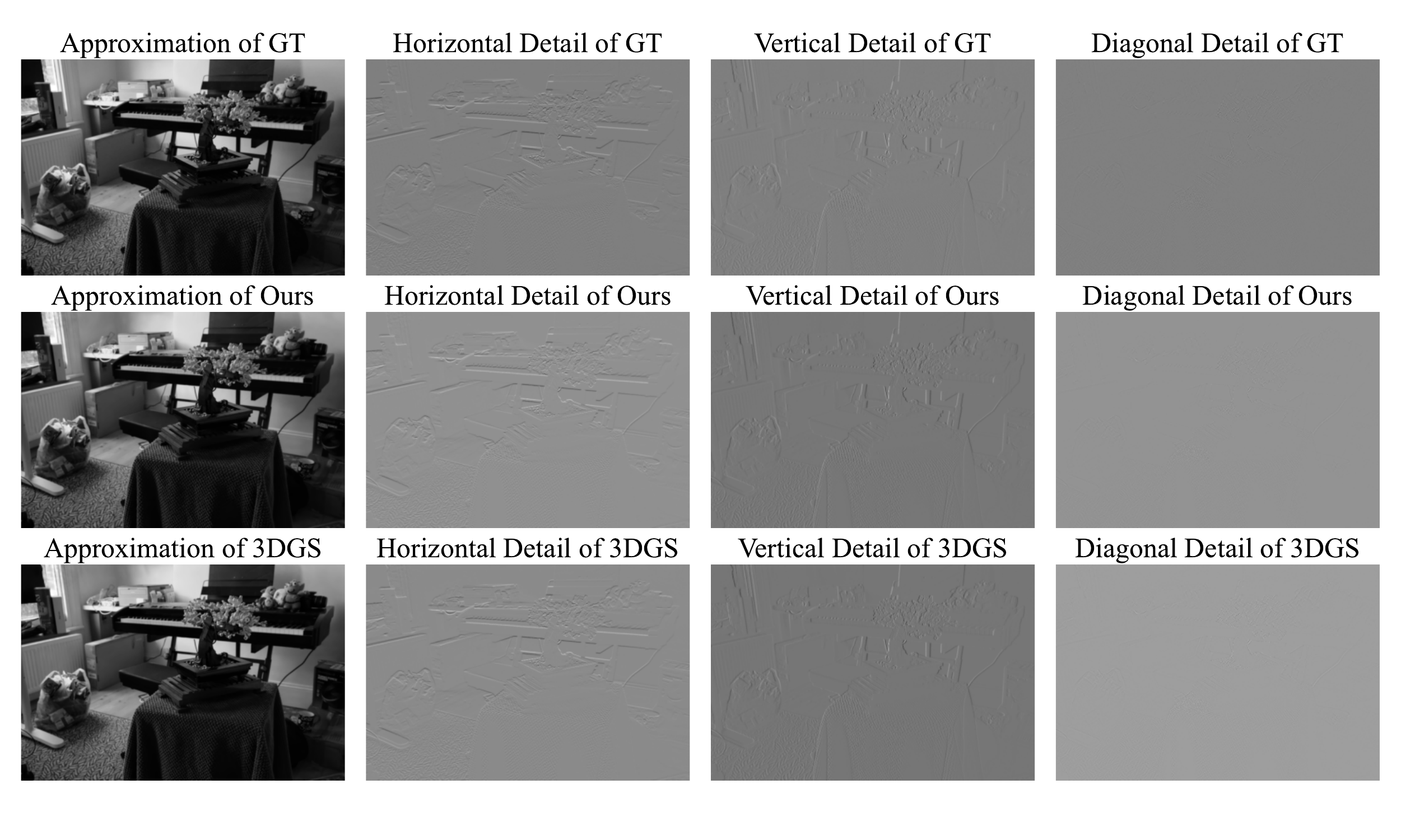}
	\caption{We perform wavelet transform on the image to visualize the boundary information of the rendered results. The horizontal and vertical detail parts after wavelet transform respectively display the boundary information in the horizontal and vertical directions of the rendered image. Obviously, our method has clearer and sharper boundaries, richer details, and low-frequency parts that are closer to real images compared to 3DGS.}
	\label{wavelet}
\end{figure*}

Our GDC enforces multi-level gradient consistency through an $\ell_1$-norm penalty:
\begin{equation}\label{eq:gdc}
    \mathcal{L}_{\text{grad}} 
    \;=\; \frac{1}{\,(H-1)(W-1)\,} 
    \sum_{i=1}^{H-1} \sum_{j=1}^{W-1} 
    \Bigl\lVert \mathbf{g}_{i,j}^\text{pred} - 
    \mathbf{g}_{i,j}^\text{gt}\Bigr\rVert_1.
\end{equation}
We exclude boundary indices where $i=H$ or $j=W$ to avoid index out-of-bounds. By minimizing the discrepancy in horizontal and vertical image gradients, the network is guided to reproduce sharp changes, such as edges, corners, and textures. We present one of the cases using wavelet transform for clearer visualization in figure~\ref{wavelet}. Decompose the boundary into horizontal and vertical directions for visual observation, and compare the low-frequency and diagonal high-frequency information. The GDC particularly enhances LH/HL band reconstruction where traditional losses fail to preserve edge orientation statistics.

This geometric regularization complements the adaptive weighting strategy through dual mechanisms: 1) Edge sharpening, direct gradient matching prevents boundary blurring in high-curvature regions. 2) Texture preservation, high-frequency gradient alignment maintains stochastic texture patterns. 3) Scale awareness, wavelet analysis reveals improved recovery across frequency subbands.

\subsection{Overall Loss Framework}

\subsubsection{Composite Objective Function}
Our complete optimization objective synthesizes three complementary mechanisms through carefully calibrated coupling coefficients:
\begin{align}
    \mathcal{L}
    &= \lambda_1 \,\mathcal{L}_{\text{w}}
       \;+\;\lambda_2\mathcal{L}_{\text{D-SSIM}}
       \;+\;\lambda_3\,\mathcal{L}_{\text{grad}},
\end{align}
where the trilinear weighting scheme $\lambda_1,\lambda_2,\lambda_3$ governs:
\begin{itemize}
    \item Local Error Correction: $\mathcal{L}_{\text{w}}$'s spatial adaptation for photometric accuracy
    \item Structural Coherence: $\mathcal{L}_{\text{D-SSIM}}$'s windowed similarity preservation
    \item Geometric Fidelity: $\mathcal{L}_{\text{grad}}$'s edge orientation constraints
\end{itemize}


Here, $\lambda_1$, $\lambda_2$, and $\lambda_3$ balance the weighted reconstruction term, SSIM, and gradient preservation. We found in practice that $\lambda_3$ can be gradually increased during training if edges remain overly smoothed.

\subsubsection{Multi-Sample Anti-Aliasing Integration}
When computing $I_\text{pred}$ at each iteration, we replace the standard pixel evaluation in Eq.~\eqref{eq:alpha_compositing} with the MSAA-based process:
\begin{equation}
    I_\text{pred}(i,j)
    \;=\; \mathbf{C}^\text{MSAA}(\mathbf{u}),
\end{equation}
where $\mathbf{u} = (j, i)$ in pixel coordinates and $\mathbf{C}^\text{MSAA}(\cdot)$ is defined as above. Hence, $I_\text{pred}$ is effectively the averaged color across multiple subpixel offsets.

By combining MSAA to reduce aliasing, the Adaptive Weighting Strategy to target high-error regions, and the Gradient Difference Constraints to retain edges, our method alleviates common pitfalls in 3D Gaussian Splatting. This leads to both more faithful local detail reconstruction and improved global rendering quality in complex scenes.

\section{Experiments} \label{sec4}
\subsection{Datasets and Implementation Details}
\subsubsection{Datasets} 
For training and testing, we followed the dataset settings of 3DGS~\cite{kerbl20233d} and conducted experiments on a total of 13 real-world images. Specifically, we evaluated our method on all nine scenarios of the Mip-NeRF360~\cite{barron2021mip} dataset, two scenarios of the Tanks\&Templates dataset~\cite{knapitsch2017tanks}, and two scenarios of the Deep Blending dataset~\cite{hedman2018deep}. The selected scene presents a variety of styles, ranging from bounded indoor environments to unbounded outdoor environments. The resolution of all images involved is also the same as in 3DGS. All scenes were trained 30000 iterations on a single RTX 4090 GPU, with hyperparameters consistent with 3DGS. Evaluate the performance of the 3DGS model and our proposed model on all datasets, including quantitative and qualitative analysis.

\begin{table*}[ht]
\centering
\renewcommand{\arraystretch}{0.8}
\caption{Quantitative comparison on three datasets. 
SSIM$\uparrow$ and PSNR$\uparrow$ are higher-the-better; 
LPIPS$\downarrow$ is lower-the-better. For fair comparison and to balance the trade-off between overall quality and memory consumption, we trained these datasets with the same settings as 3DGS. All methods use the same training data for training. The \colorbox{red!25}{best score}, \colorbox{orange!25}{second best score}, and \colorbox{yellow!30}{third best score} are red, orange, and yellow, respectively.}
\label{tab:results}
\setlength{\tabcolsep}{5pt} 
\resizebox{\textwidth}{!}{
\begin{tabular}{
l
|ccc
|ccc
|ccc
}
\toprule
\textbf{Datasets}& \multicolumn{3}{c}{\textbf{Mip-NeRF360}} 
& \multicolumn{3}{|c}{\textbf{Tanks\&Temples}} 
& \multicolumn{3}{|c}{\textbf{Deep Blending}} 
\\
\cmidrule(lr){1-2} \cmidrule(lr){2-4} \cmidrule(lr){5-7} \cmidrule(lr){8-10}
\textbf{Methods} 
& SSIM$\uparrow$ & PSNR$\uparrow$ & LPIPS$\downarrow$
& SSIM$\uparrow$ & PSNR$\uparrow$ & LPIPS$\downarrow$
& SSIM$\uparrow$ & PSNR$\uparrow$ & LPIPS$\downarrow$
\\
\midrule
Plenoxels
&0.626  &23.08  &0.463 
&0.719  &21.08  &0.379
&0.795  &23.06  &0.510 
\\
INGP
&0.671  &25.30  &0.371 
&0.723  &21.72  &0.330 
&0.797  &23.62  &0.423 
\\
{Mip-NeRF360}
& {0.792} & {27.69} & {0.237}
& {0.759} & {22.22} & {0.257}
& {0.901} & {29.40} & {0.245}
\\
{3DGS}
&{0.815}      &{27.21}       &{0.214} 
&{0.841}       &{23.14}      &{0.183}
&\ccy{0.903} &{29.41} &{0.243}
\\
Pixel-GS
&{0.823}      & \cco{27.67}       & {0.193} 
&\ccy{0.856}       & \ccy{23.74}      & \cco{0.151}
&{0.896} & {28.91} & {0.248}
\\
AbsGS
&\ccy{0.820}      & {27.49}       & \ccy{0.191} 
&{0.853}       & {23.73}      & {0.162}
&{0.902} & \cco{29.67} & \cco{0.236}
\\
\hline
\textbf{Ours(3DGS)}
& {0.819} & \ccy{27.62} & {0.207}
& {0.851} & \ccr{23.79} & {0.165}        	
& {0.900} & \ccy{29.52} & {0.246}
\\
\textbf{Ours(Pixel-GS)}
&\ccr{0.830} & \ccr{27.68} & \cco{0.189} 
&\ccr{0.865} & \cco{23.75} & \ccr{0.145}
&\ccr{0.907} & {28.92} & \ccy{0.242}
\\
\textbf{Ours(AbsGS)}
&\cco{0.829} & {27.51} & \ccr{0.188} 
&\cco{0.862} & \ccy{23.74} & \ccy{0.157}
&\cco{0.906} & \ccr{29.69} & \ccr{0.230}
\\

\bottomrule
\end{tabular}}
\end{table*}

\subsubsection{Implementation} 
In our rendering reconstruction framework, we integrate three enhancements. First, to reduce aliasing and jagged edges, $4\times$ MSAA is employed at the rendering stage to collect multiple samples per pixel, which are then fused either at the input layer or within intermediate feature fusion modules. Second, our Adaptive Weighting Strategy dynamically modulates the loss weights based on pixel or region-level reconstruction errors, thus emphasizing challenging regions with higher errors. Finally, to preserve high-frequency details and sharp boundaries, we apply Gradient Difference Constraints using a gradient operator on both the predicted and ground-truth images, and incorporate the resulting discrepancy into the overall loss function. We use the Adam optimizer for training in the Pytorch framework\cite{paszke2019pytorch}, integrate MSAA into the rasterization of 3DGS, and set the $\lambda_1,\lambda_2,\lambda_3$ to $0.8$, $0.2$, and $0.1$.

\begin{table*}[ht]
\centering
\caption{Quantitative comparison on Mip-NeRF360 dataset. 
SSIM$\uparrow$ and PSNR$\uparrow$ are higher-the-better, using $+$ to indicate the improvement of our method compared to 3DGS; 
LPIPS$\downarrow$ is lower-the-better, using $-$ to represent the improvement of our method compared to 3DGS. For fair comparison and to balance the trade-off between overall quality and memory consumption, we trained these datasets with the same settings as 3DGS. All methods use the same training data for training. The \colorbox{red!25}{best score} are red.}
\label{Mip-NeRF360}
\renewcommand{\arraystretch}{1.5}
\resizebox{\textwidth}{!}{\setlength{\tabcolsep}{1mm}{
\begin{tabular}{
l
|ccc
|ccc
|ccc
}
\toprule
\textbf{Datasets}& \multicolumn{3}{c}{\textbf{bicycle}} 
& \multicolumn{3}{|c}{\textbf{bonsai}} 
& \multicolumn{3}{|c}{\textbf{counter}} 
\\
\cmidrule(lr){1-2} \cmidrule(lr){2-4} \cmidrule(lr){5-7} \cmidrule(lr){8-10}
\textbf{Methods} 
& SSIM$\uparrow$ & PSNR$\uparrow$ & LPIPS$\downarrow$
& SSIM$\uparrow$ & PSNR$\uparrow$ & LPIPS$\downarrow$
& SSIM$\uparrow$ & PSNR$\uparrow$ & LPIPS$\downarrow$
\\
\midrule
{3DGS}
&0.745 &25.08 &0.245
&0.947 &32.36 &0.179
&0.915 &29.11 &0.182
\\       	
\textbf{Ours}
&\ccr{0.758 (+0.013)} &\ccr{25.23 (+0.15)} &\ccr{0.222 (-0.023)}
&\ccr{0.949 (+0.002)} &\ccr{32.44 (+0.08)} &\ccr{0.173 (-0.006)}
&\ccr{0.918 (+0.003)} &\ccr{29.12 (+0.01)} &\ccr{0.175 (-0.007)}
\\
\hline
\textbf{Datasets}& \multicolumn{3}{c}{\textbf{flowers}} 
& \multicolumn{3}{|c}{\textbf{garden}} 
& \multicolumn{3}{|c}{\textbf{kitchen}} 
\\
\cmidrule(lr){1-2} \cmidrule(lr){2-4} \cmidrule(lr){5-7} \cmidrule(lr){8-10}
\textbf{Methods} 
& SSIM$\uparrow$ & PSNR$\uparrow$ & LPIPS$\downarrow$
& SSIM$\uparrow$ & PSNR$\uparrow$ & LPIPS$\downarrow$
& SSIM$\uparrow$ & PSNR$\uparrow$ & LPIPS$\downarrow$
\\
\hline
{3DGS}
&0.589      &21.40      &0.358 
&0.856      &27.28      &0.122
&0.931      &31.32      &0.116
\\      	
\textbf{Ours}
&\ccr{0.606 (+0.017)} &\ccr{21.69 (+0.29)} &\ccr{0.339 (-0.019)}
&\ccr{0.863 (+0.007)} &\ccr{27.41 (+0.13)} &\ccr{0.109 (-0.013)}
&\ccr{0.934 (+0.003)} &\ccr{31.45 (+0.13)} &\ccr{0.111 (-0.005)}
\\
\hline
\textbf{Datasets}& \multicolumn{3}{c}{\textbf{room}} 
& \multicolumn{3}{|c}{\textbf{stump}} 
& \multicolumn{3}{|c}{\textbf{treehill}} 
\\
\cmidrule(lr){1-2} \cmidrule(lr){2-4} \cmidrule(lr){5-7} \cmidrule(lr){8-10}
\textbf{Methods} 
& SSIM$\uparrow$ & PSNR$\uparrow$ & LPIPS$\downarrow$
& SSIM$\uparrow$ & PSNR$\uparrow$ & LPIPS$\downarrow$
& SSIM$\uparrow$ & PSNR$\uparrow$ & LPIPS$\downarrow$
\\
\hline
{3DGS}
&0.926      &31.70      &0.196
&0.768      &26.63      &0.243
&0.635      &22.52      &0.346
\\   	
\textbf{Ours}
&\ccr{0.929 (+0.003)} &\ccr{31.78 (+0.08)} &\ccr{0.186 (-0.010)}
&\ccr{0.778 (+0.010)} &\ccr{26.81 (+0.18)} &\ccr{0.224 (-0.019)}
&\ccr{0.643 (+0.008)} &\ccr{22.63 (+0.11)} &\ccr{0.327 (-0.019)}

\\
\bottomrule
\end{tabular}}}
\end{table*}

\begin{figure}[H]
	\centering
	\includegraphics[width=\textwidth]{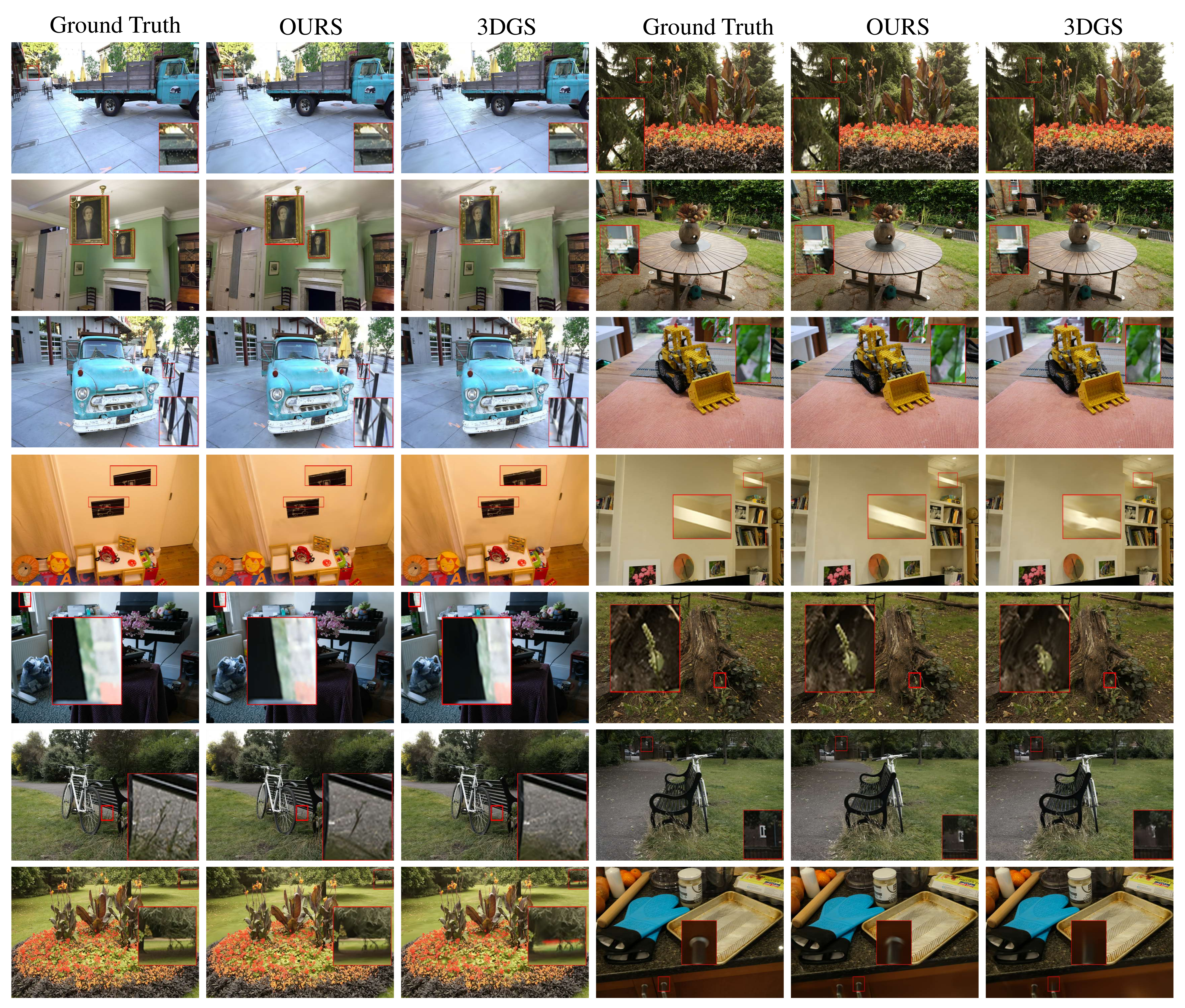}
	\caption{Qualitative comparison between our proposed method and state-of-the-art methods in novel view synthesis. The comparison was conducted in multiple scenes with different styles, including multiple scenes from Mip-NeRF360, Tank\&Template's trains and trucks, and Deep Blending's playroom. The GT in the figure represents the real image used for reference, and we will zoom in on the clearly different details for easier viewing. From these results, it can be seen that our method achieves excellent image rendering with fewer artifacts and richer details.}
	\label{quantitative}
\end{figure}

\subsection{Results Analysis}
\begin{figure}[ht]
	\centering
	\includegraphics[width=0.9\textwidth]{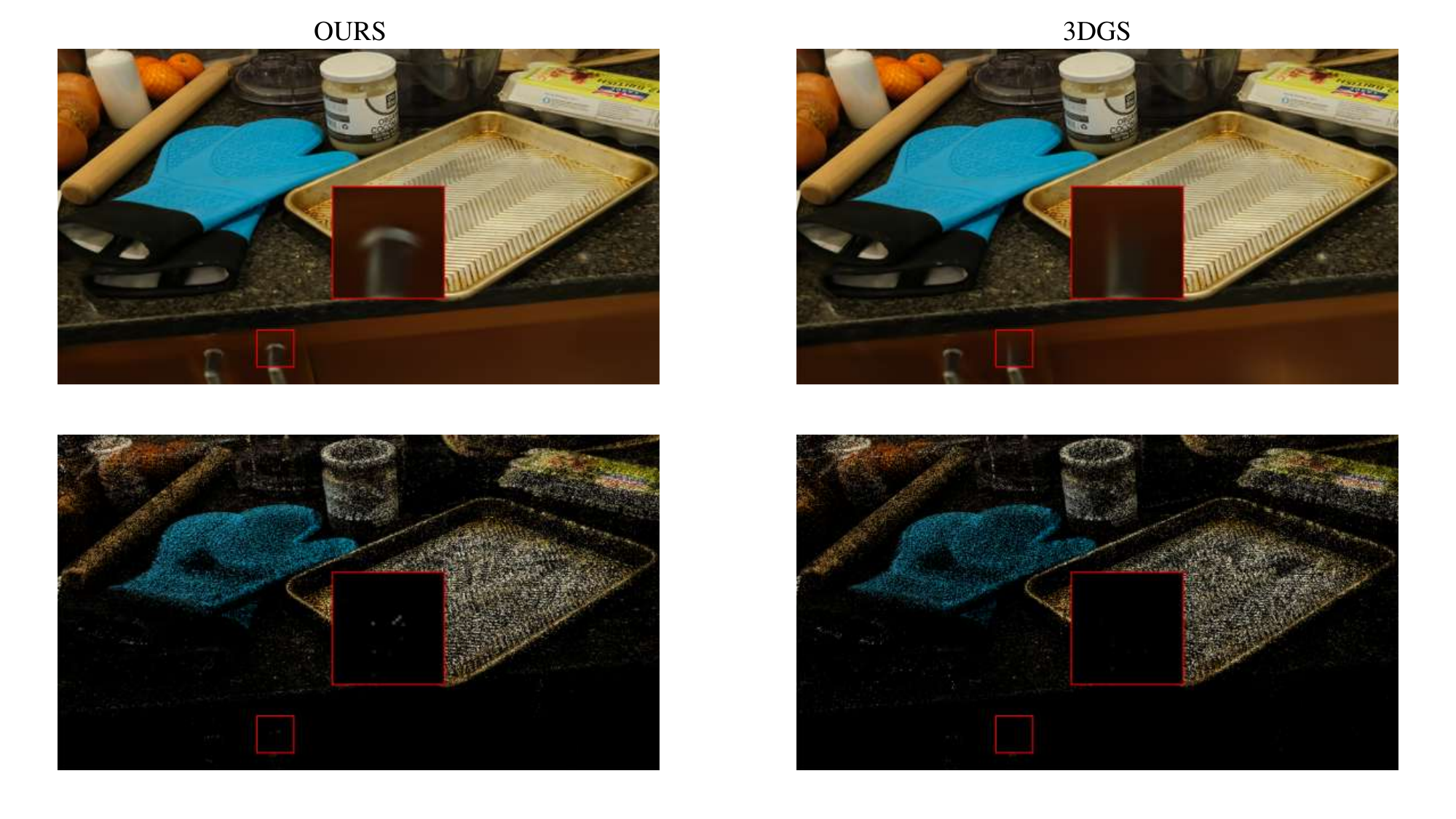}
	\caption{Comparison diagram of point cloud effect. Our method outperforms 3DGS in reconstructing detailed point clouds, especially in blurred or textureless areas, enhancing local rendering quality.}
	\label{detail}
\end{figure}

\begin{figure}[ht]
	\centering
	\includegraphics[width=0.9\textwidth]{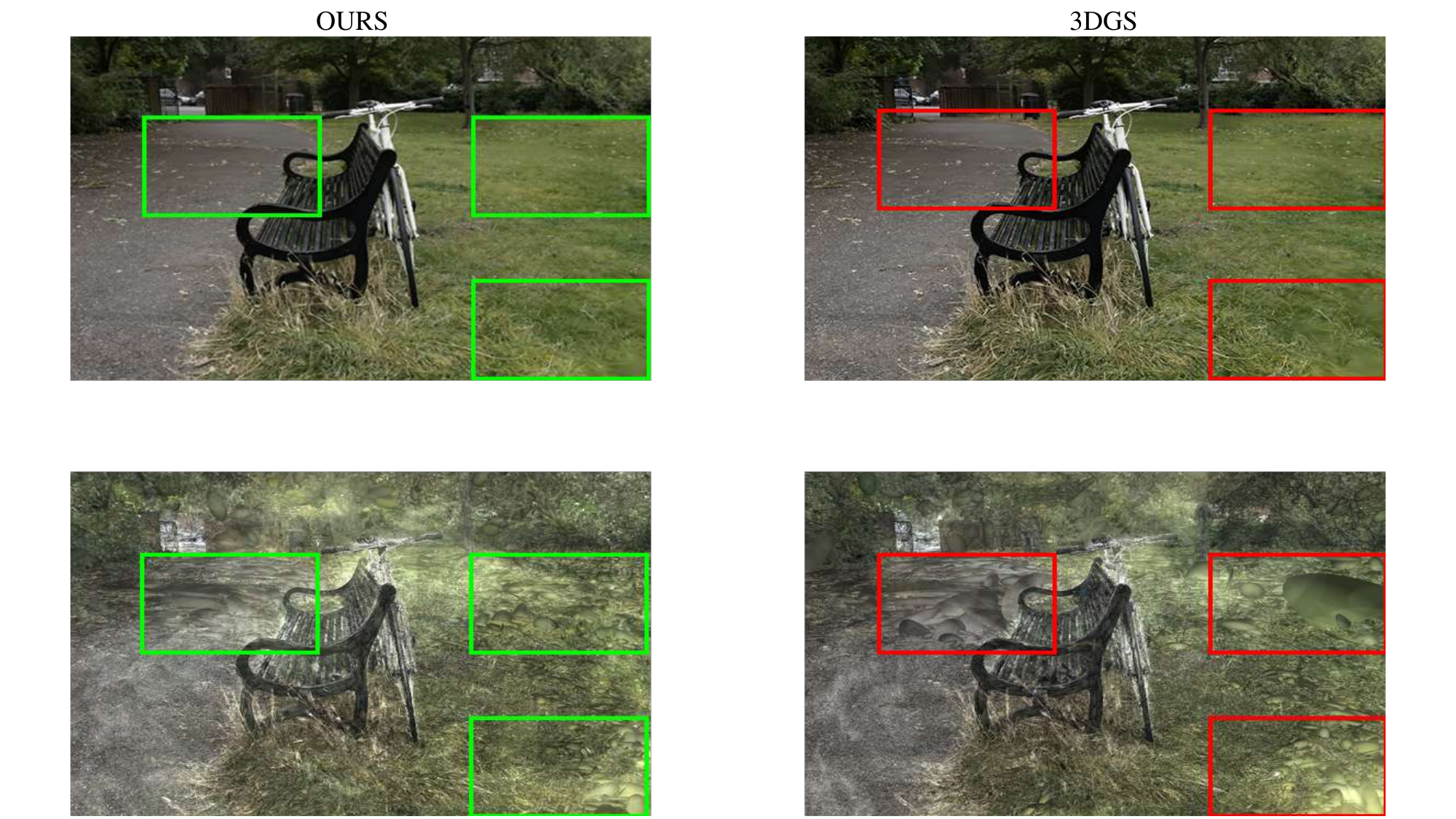}
	\caption{Gaussian ellipsoid coverage comparison diagram. Our method resolves 3DGS's blurry rendering by decomposing large Gaussian ellipsoids into finer, detail-rich components for enhanced reconstruction.}
	\label{detail_P}
\end{figure}

We compared our approach with 3DGS as well as other view-generation methods—including MipNerf360~\cite{barron2021mip}, InstantNGP~\cite{muller2022instant}, Plenoxels~\cite{fridovich2022plenoxels}, Mip-NeRF360~\cite{barron2022mip}, 3DGS~\cite{kerbl20233d}, AbsGS~\cite{ye2024absgs} and Pixel-GS~\cite{zhang2024pixelgs}—across 14 scenarios from three datasets. To ensure a fair comparison and balance between memory usage and performance, we employed a similar number of Gaussians as in 3DGS. All methods were trained using the same data and hardware configurations. Table \ref{tab:results} presents the quantitative metrics for all methods, and Table \ref{Mip-NeRF360} details results for the nine scenarios in MipNerf360.
As shown in these analyses, our method consistently outperforms state-of-the-art approaches in terms of PSNR, SSIM, and LPIPS for all real-world scenarios. Moreover, the quantitative results in Figure \ref{quantitative} further illustrate our superior rendering quality, highlighting the robustness of our approach in diverse environments. In addition, we use point clouds and Gaussian ellipsoids to demonstrate the effectiveness of the proposed dual constraints in Figure~\ref{detail} and Figure~\ref{detail_P}.

\begin{figure}[ht]
	\centering
	\includegraphics[width=\linewidth]{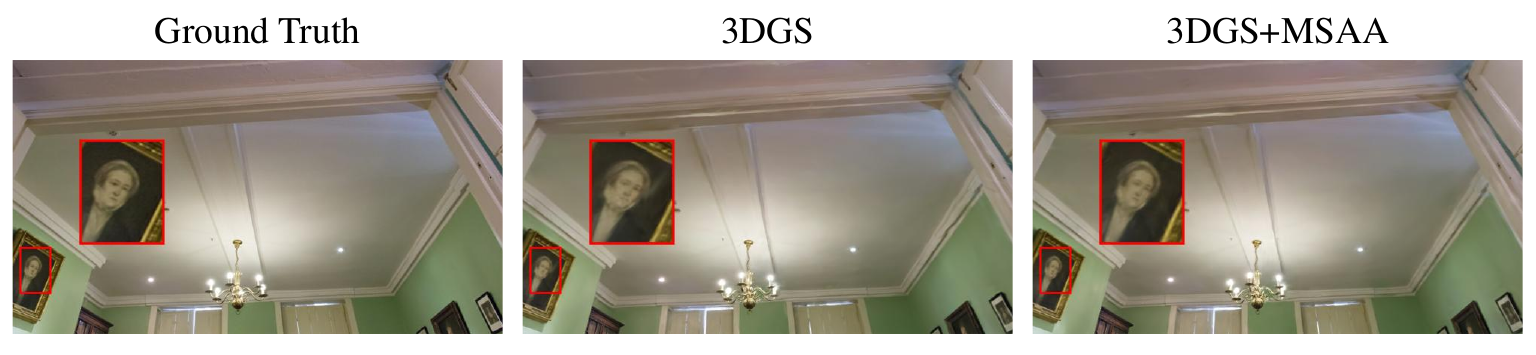}
	\caption{The results of the MSAA ablation experiment. 3DGS generates some non-existent parts when reconstructing and rendering portraits, which reduces the clarity of the rendering results. After adding the MSAA method, we effectively distinguished the areas of the portrait, blank space, and frame during reconstruction and rendering, making the content of each part clearer.}
	\label{MSAA}
\end{figure}

\subsection{Ablation Studies}
To investigate the individual contributions of our proposed improvement modules, we conducted ablation experiments on MSAA, Adaptive Weighting Strategy (AWS), and Gradient Difference Constraints (GDC). Specifically, we sequentially removed or selectively retained each module from the full model configuration, then evaluated and compared each variant using standard reconstruction metrics.

\subsubsection{Multisample Anti-Aliasing}
We first investigate how our proposed MSAA affects local blurring in Gaussian distributions. Based on 3DGS, we trained a model 3DGS+MSAA, which executes the MSAA strategy during the rasterization stage. As shown in the figure \ref{MSAA}, the 3DGS+MSAA model is clearer in local details, MSAA effectively alleviates jagged edges and aliasing distortion during rendering, making the network less prone to noise and defocusing when dealing with sharp edges or high-frequency textures.

\begin{table*}[ht]
    \centering
    \renewcommand{\arraystretch}{0.4}
    \caption{Conduct ablation experiments on the Mip-NeRF360 and Tank\&Temples datasets. SSIM$\uparrow$ and PSNR$\uparrow$ are higher-the-better; LPIPS$\downarrow$ is lower-the-better. The \colorbox{red!25}{best score}, \colorbox{orange!25}{second best score} are red, orange, respectively.}
    \label{Ablation}
    \resizebox{\linewidth}{!}{\setlength{\tabcolsep}{1mm}{
    \begin{tabular}{l|ccc|ccc|ccc}
    \toprule
    \multirow{2}{*}{\textbf{Methods}} 
    & \multicolumn{3}{c|}{\textbf{Mip-NeRF360}} 
    & \multicolumn{3}{c|}{\textbf{Tank\&Temple}}
    & \multicolumn{3}{c}{\textbf{Deep Blending}}
    \\
    \cmidrule{2-10}
    & SSIM$\uparrow$ & PSNR$\uparrow$ & LPIPS$\downarrow$
    & SSIM$\uparrow$ & PSNR$\uparrow$ & LPIPS$\downarrow$ 
    & SSIM$\uparrow$ & PSNR$\uparrow$ & LPIPS$\downarrow$
    \\
    \midrule
    3DGS        &0.815      & 27.21       & \cco{0.214}      &0.841       & 23.14      & 0.183     &\ccr{0.903} &29.41 &\ccr{0.243} \\
    3DGS+MSAA     &0.812       &\cco{27.50}       &0.221       &0.844       &\cco{23.69}       &0.178 &0.899 &\cco{29.48} &0.246     \\
    3DGS+AWS+GDC   &\cco{0.}816      &27.45      &0.210       &\cco{0.846}       &23.66       &\cco{0.170} &0.901 &29.44 &\cco{0.244 }     \\
    Ours  & \ccr{0.819} & \ccr{27.62} & \ccr{0.207}      &  \ccr{0.851}     & \ccr{23.79}      & \ccr{0.165}   &\cco{0.900} &\ccr{29.52} &0.246  \\
    \bottomrule
    \end{tabular}}}
\end{table*}

\subsubsection{Adaptive Weighting Strategy \& Gradient Difference Constraints}
To validate the effectiveness of our proposed constraint optimization framework, we conduct comprehensive ablation studies by training a hybrid model, 3DGS+MSAA+AWS+GDC, which integrates MSAA with our adaptive constraint optimization strategies. Specifically, our method outperforms both 3DGS+MSAA and 3DGS+AWS+GDC across key metrics, demonstrating the synergistic benefits of combining adaptive weighting and gradient-domain regularization.
The AWS dynamically adjusts loss weights for challenging regions based on per-pixel reconstruction errors. This mechanism prioritizes under-optimized areas during training, significantly enhancing the recovery of fine texture details while mitigating visual artifacts like local blurring and over-smoothing caused by uneven optimization focus. Complementing AWS, the GDC constraint enforces coherence between adjacent pixels in the gradient space, effectively suppressing excessive smoothing while preserving high-frequency features and sharp structural edges.
As quantitatively verified in Table~\ref{Ablation}, our full framework achieves superior novel view synthesis quality compared to 3DGS+MSAA,
while also surpassing the standalone 3DGS+AWS+GDC configuration. These results confirm 
that our constraint optimization paradigm not only resolves the limitations of MSAA in
handling complex textures but also provides a more balanced and robust optimization 
landscape compared to isolated constraint strategies.

\section{Conclusion}
In summary, our approach leverages a pixel-level weighted constraint, MSAA and gradient difference constraints to address the challenges of reconstructing fine local details and mitigating aliasing in 3D Gaussian splatting. By assigning adaptive weights based on pixel gradients, our method effectively focuses on areas with higher reconstruction difficulty, while MSAA alleviates jagged boundaries. Furthermore, incorporating gradient difference constraints preserves high-frequency signals and sharp edges. As a result, our method demonstrates remarkable improvements over naive splatting and achieves state-of-the-art performance in local detail reconstruction.

\section*{Declarations}


\begin{itemize}
\item Conflict of interest/Competing interests (check journal-specific guidelines for which heading to use): The authors have no competing interests to declare that are relevant to the content of this article.
\item Data availability: The data used in this article has been declared and cited in the experimental section~\ref{sec4}.
\end{itemize}

\bibliography{sn-bibliography}

\end{document}